\documentclass[11pt]{article}

\usepackage[preprint]{acl}

\usepackage{times}
\usepackage{latexsym}
\usepackage[T1]{fontenc}
\usepackage[utf8]{inputenc}
\usepackage{microtype}
\usepackage{inconsolata}
\usepackage{graphicx}
\usepackage{natbib}
\usepackage{booktabs}

\title{Enhancing Zero-Shot Domain Transfer in Low-Resource Sentiment Analysis: A Case Study in Marathi}
 
\author{
Tanvi Somani$^{1,3}$, Aryan Babare$^{1,3}$, Samyak Bora$^{1,3}$, Raviraj Joshi$^{2,3}$ \\
$^{1}$Pune Institute of Computer Technology (PICT), Pune, Maharashtra, India \\
$^{2}$Indian Institute of Technology Madras, Chennai, Tamil Nadu, India \\
$^{3}$L3Cube Labs, Pune \\
\texttt{\{tanvisomani0216, aryanbabare1, samyakkiranbora\}@gmail.com} \\
\texttt{ravirajoshi@gmail.com}
}

\begin{document}
\maketitle
\begin{abstract}
Sentiment analysis is a key task in Natural Language Processing that determines the polarity of opinions in text. For low-resource languages like Marathi—where limited data prevents the direct use of standard statistical methods \cite{magueresse2020low}—models are often trained on single-domain datasets. While this yields strong in-domain performance, it leads to poor generalization to unseen domains. Recent studies highlight the need for domain-robust systems that can handle diverse real-world content through multi-domain modeling \cite{toledo-ronen-etal-2022-multi}.
This study investigates how multi-domain training improves sentiment classification for unseen Marathi domains. Models are trained using single-domain, two-domain, and three-domain dataset combinations. Results show that two-domain training improves accuracy on unseen domains by about 8\% compared to single-domain training. Adding a third domain provides further improvement, though the gain is smaller, averaging around 2\% for some domains.
These results suggest that training with data from multiple domains helps build more robust sentiment analysis models for low-resource languages, though the benefits reduce as more domains are added.

\textbf{Keywords—} Marathi Sentiment Analysis, Out-of-Domain Generalization, Zero-Shot Learning, Multi-Domain Training, Low-Resource Languages, Natural Language Processing.

\end{abstract}

\section{Introduction}
Sentiment classification is a key task in NLP, supporting applications such as opinion mining and user feedback analysis. Emotion recognition is also widely used in domains like education and personalized web systems to adapt to user behavior \cite{kolakowska2015emotion}. With the increasing presence of regional language content online, effective sentiment analysis for languages like Marathi is needed. However, progress remains limited due to scarce annotated data and narrow domain coverage. Recent studies show that Transformer-based models outperform traditional approaches in multilingual low-resource settings \cite{aliyu2024comparative}.

A major challenge in Marathi sentiment analysis is poor generalization to unseen domains. Models trained on a single domain often suffer a sharp decline in accuracy when evaluated on different distributions, reflecting weak zero-shot capabilities. While manifold-based augmentation \cite{ng-etal-2020-ssmba} and simple strategies like EDA \cite{wei-zou-2019-eda} can improve performance on small datasets, achieving true out-of-domain robustness in low-resource languages remains a significant challenge.

Although recent transformer-based models have improved sentiment classification for Marathi, most studies emphasize performance within a fixed domain. Limited attention has been given to understanding how exposure to diverse domains during training influences model behavior on unseen domains. As a result, there is a lack of empirical analysis exploring data-driven strategies to improve cross-domain robustness in Marathi sentiment classification.

In this work, we investigate the effect of training data diversity on the generalization capability of sentiment classification models. Rather than introducing new model architectures, we focus on combining sentiment datasets from multiple domains and training models under different domain combination settings, such as two-domain and three-domain configurations. This approach allows the model to learn a broader range of linguistic patterns and sentiment cues.

Through systematic evaluation on unseen domains, this study examines how increasing domain diversity during training impacts out-of-domain and zero-shot generalization performance. The findings of this work provide valuable insights into designing more robust Marathi sentiment analysis systems that are better suited for real-world, multi-domain applications. The main contributions of this work are as follows:
\begin{itemize}
    \item This study investigates cross-domain sentiment classification in Marathi, focusing on improving generalization in low-resource settings through multi-domain training.
    \item We design and evaluate multi-domain training configurations (two-domain and three-domain) to analyze out-of-domain and zero-shot generalization on unseen domains.
    \item We demonstrate that increasing domain diversity improves model robustness, and that multi-domain training outperforms single-domain approaches for real-world applications.
\end{itemize}


\section{Related Work}

Sentiment analysis has been revolutionized by the introduction of Bidirectional Encoder Representations from Transformers (BERT), which pre-trains deep bidirectional representations by conditioning on both left and right context \cite{devlin-etal-2019-bert}. While BERT-based models achieve state-of-the-art results in language understanding, their cross-domain generalization remains a challenge for low-resource languages like Marathi.

 \citet{roccabruna2022} examined sentiment analysis across multiple sources and domains using BERT-based models. Their work evaluated several datasets collected from domains such as social media, finance, and e-commerce. The study found that models trained on data from multiple domains generally perform better when applied to unseen domains compared to models trained on a single dataset. However, the authors also observed that model performance still varies depending on the domain, and that predicting neutral sentiment is more challenging than predicting positive or negative sentiment. These findings suggest that domain diversity during training improves robustness but does not completely solve the cross-domain generalization problem.

For Marathi, recent research has focused on creating datasets and language resources to support sentiment analysis \cite{kulkarni2021l3cubemahasent}. \citet{pingle2023mahasent} introduced the L3Cube-MahaSent-MD dataset, which contains sentiment-labeled data collected from four domains: movie reviews, political tweets, general tweets, and television subtitles. The dataset includes approximately 60,000 manually annotated samples with three sentiment labels: positive, negative, and neutral. The authors evaluated several transformer models including mBERT, MuRIL, IndicBERT, and MahaBERT in both in-domain and cross-domain settings. Their results highlight the usefulness of multi-domain datasets for studying generalization in Marathi sentiment analysis.

Similarly, \citet{joshi2022mahacorpus} introduced L3Cube-MahaCorpus, a large collection of Marathi text gathered from various online sources. Using this corpus, several Marathi language models such as MahaBERT, MahaALBERT, and MahaRoBERTa were trained. These models help capture the linguistic structure and semantics of Marathi more effectively and have shown improved results on tasks such as text classification and named entity recognition. For low-resource languages like Marathi, the availability of such datasets and pretrained models plays an important role in advancing NLP research.

Researchers have also explored methods to improve sentiment model robustness when training data is limited. In another study, \citet{pingle2023augmentation} investigated several data augmentation techniques for Marathi sentiment analysis. Their experiments included approaches such as back-translation, a method shown to be highly effective in generating large-scale pseudo-parallel corpora for improving model accuracy in context-aware settings \cite{sugiyama-yoshinaga-2019-data}. While these studies focus on expanding datasets through synthetic means, our work explores the impact of combining diverse, human-annotated domains to achieve similar robustness.

Although these works provide useful datasets, pretrained models, and training strategies, relatively little research has directly examined how training on multiple domains affects model performance on unseen domains in Marathi. Understanding this relationship is important for developing sentiment classification models that can perform reliably in real-world scenarios where text data originates from diverse sources.

\section{Dataset}
In this section, we describe the MAHASENT-MD multi-domain Marathi sentiment dataset used for training and evaluation.
\subsection{MAHASENT-MD Dataset}

To evaluate cross-domain generalization in Marathi sentiment classification, we employ a multi-domain corpus referred to as \textbf{MAHASENT-MD} (Marathi Sentiment Multi-Domain). The dataset consists of sentiment-annotated textual data collected from four heterogeneous domains: \textit{Movie Reviews}, \textit{Political Tweets}, \textit{Subtitles}, and \textit{Generic Tweets}. Each instance in the dataset is manually annotated with one of three sentiment labels: \textit{Positive}, \textit{Negative}, or \textit{Neutral}.

Let the set of domains be defined as:

\begin{equation}
D = \{D_1, D_2, D_3, D_4\}
\end{equation}

where $D_1$ corresponds to Movie Reviews, $D_2$ corresponds to Political Tweets, $D_3$ represents Subtitles, and $D_4$ represents Generic Tweets.

Each domain represents a distinct textual distribution $P_i(x,y)$, characterized by variations in vocabulary usage, contextual patterns, writing style, and sentiment expression. These domain-specific characteristics introduce distributional differences that affect the transferability of sentiment classification models.

For example, the Subtitles domain ($D_3$) typically contains informal language, abbreviations, hashtags, and emotive expressions, resulting in high lexical variability. In contrast, the Generic Tweets domain ($D_4$) covers a broader range of topics and exhibits mixed linguistic formality with diverse contextual structures.

Such inter-domain variations introduce distributional shifts that challenge the ability of models to generalize beyond their training domains. By incorporating multiple heterogeneous domains, the MAHASENT-MD dataset provides an appropriate framework for investigating out-of-domain robustness and zero-shot transfer capability in Marathi sentiment classification.
Table~\ref{tab:dataset_stats} presents the detailed statistics and data distribution across different domains in the MAHASENT-MD dataset.


\begin{table*}[t]
\centering
\caption{Domain characteristics of the MahaSent-MD Dataset.}
\label{tab:dataset_stats}
\vspace{0.3cm}
\renewcommand{\arraystretch}{1.2}
\setlength{\tabcolsep}{12pt}
\begin{tabular}{lccc}
\hline
\textbf{Domain} & \textbf{Approx Size} & \textbf{Text Nature} & \textbf{Avg Length} \\
\hline
Political Tweets (PT) & $\sim$16K & Formal, opinion-driven, topic-focused & 26.8658 \\
Generic Tweets (GT) & $\sim$15K & Informal, diverse, user-generated & 10.4528 \\
Subtitles (ST) & $\sim$15K & Conversational, short utterances & 6.3753 \\
Movie Reviews (MR) & $\sim$15K & Descriptive, expressive, sentiment-rich & 12.6323 \\
\hline
\end{tabular}
\end{table*}

\section{Methodology}

\subsection{Multi-Domain Configuration Strategy}

\begin{figure*}[t]
  \centering
  \includegraphics[width=\textwidth]{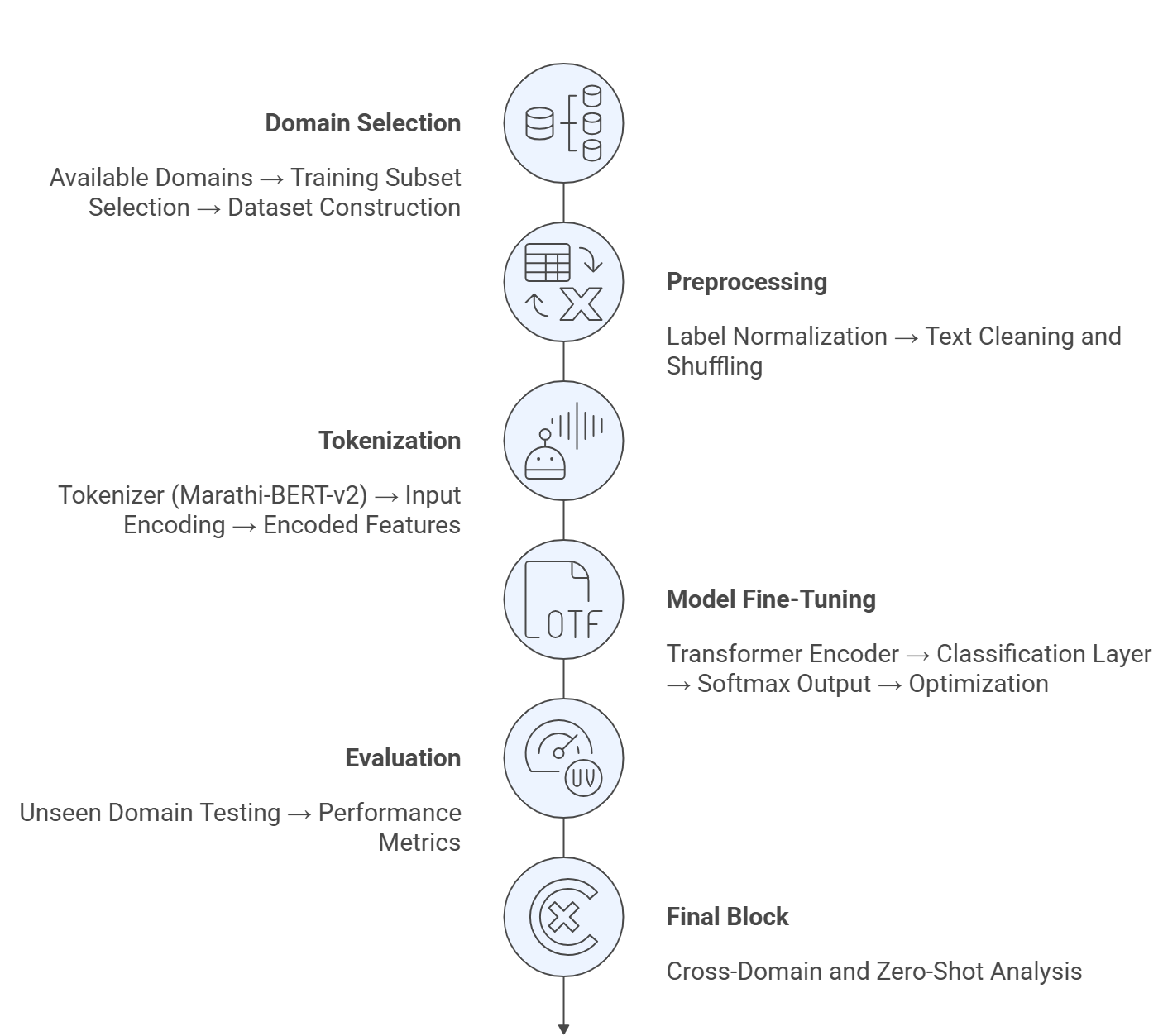}
  \caption{Multi-Domain Training and Evaluation Pipeline.}
  \label{fig:experiments}
\end{figure*}

To systematically examine the effect of domain diversity on sentiment classification performance, we construct multiple training configurations using the MAHASENT-MD corpus. The proposed experimental framework progressively expands the training distribution by varying the number of contributing domains while maintaining the same model architecture across all configurations. This design ensures that performance variations can be attributed solely to differences in training data composition. An overview of the framework is illustrated in Fig.~\ref{fig:experiments}.

\subsubsection{Single-Domain Configuration}

In the single-domain configuration, the model is trained using samples from only one domain $D_i$ and evaluated on the remaining unseen domains. For instance, a model trained solely on the Subtitles domain is evaluated on Generic Tweets, Movie Reviews, and Political Tweets. This configuration represents the baseline scenario and provides a lower-bound estimate of out-of-domain generalization. Since the model is exposed to a limited distribution during training, it typically exhibits strong domain bias and reduced performance when evaluated on heterogeneous domains.

\subsubsection{Two-Domain Configuration}

In the two-domain configuration, datasets from two heterogeneous domains are combined to construct the training corpus. Formally, the training set is defined as:

\begin{equation}
D_{train} = D_i \cup D_j
\end{equation}

where $D_i$ and $D_j$ represent two distinct domains. By learning sentiment patterns from multiple textual distributions, the model is exposed to diverse lexical, stylistic, and contextual variations. For example, combining Subtitles with Generic Tweets allows the model to observe both informal conversational text and broader topical discussions. Evaluation is then performed on a third unseen domain to assess the model's cross-domain transfer capability.

\subsubsection{Three-Domain Configuration}

The three-domain configuration further expands the training distribution by combining three domains while reserving the remaining domain exclusively for evaluation. In this setting, the training set can be expressed as:

\begin{equation}
D_{train} = D_i \cup D_j \cup D_k
\end{equation}

where $D_i$, $D_j$, and $D_k$ denote three distinct domains. For example, the model may be trained on Movie Reviews, Political Tweets, and Subtitles while being evaluated on Generic Tweets. This setup simulates a zero-shot evaluation scenario in which the model encounters a previously unseen domain during inference.

\subsubsection{Discussion}

By progressively increasing the number of training domains, the proposed framework enables systematic analysis of how exposure to heterogeneous linguistic distributions affects model generalization. This controlled experimental design isolates the influence of domain diversity from architectural factors and provides insights into the relationship between domain coverage and cross-domain sentiment classification performance.

\subsection{Model Architecture}

We utilize Transformer-based \cite{vaswani2017attention} pretrained language models for sentiment classification. These architectures, including optimized variants like ALBERT \cite{lan2019albert}, use parameter-reduction techniques to improve training efficiency and scaling. The encoder generates contextualized representations, capturing semantic dependencies within Marathi text, while the final hidden state is passed through a softmax layer for polarity prediction.

The same architectural configuration is maintained across all experimental settings to ensure that observed performance variations are attributable solely to differences in training data composition.



\subsection{Training Procedure}

For each experimental configuration, sentiment classification models are fine-tuned using domain-combined datasets derived from the MAHASENT-MD corpus. The framework supports training over arbitrary subsets of domains in order to analyze the effect of domain diversity on model generalization. Let 

\begin{equation}
D = \{D_1, D_2, \dots, D_m\}
\end{equation}

denote the complete set of available domains. For a selected subset $S \subseteq D$, the training dataset is constructed by combining samples from all domains belonging to $S$. Similarly, the validation dataset is created using the corresponding validation splits of the selected domains. The merged datasets are randomly shuffled prior to training in order to eliminate ordering bias during the optimization process.

The original sentiment labels in the dataset are encoded as

\begin{equation}
Y = \{-1, 0, 1\}
\end{equation}

representing negative, neutral, and positive sentiment, respectively. For compatibility with standard multi-class classification frameworks, the labels are transformed using the mapping

\begin{equation}
y' = y + 1
\end{equation}

which results in the label space

\begin{equation}
Y' = \{0,1,2\}.
\end{equation}

This transformation ensures proper indexing for the softmax output layer during training.







latex
\begin{figure*}[p]
\centering
\vspace*{\fill}

\includegraphics[width=0.5\textwidth]{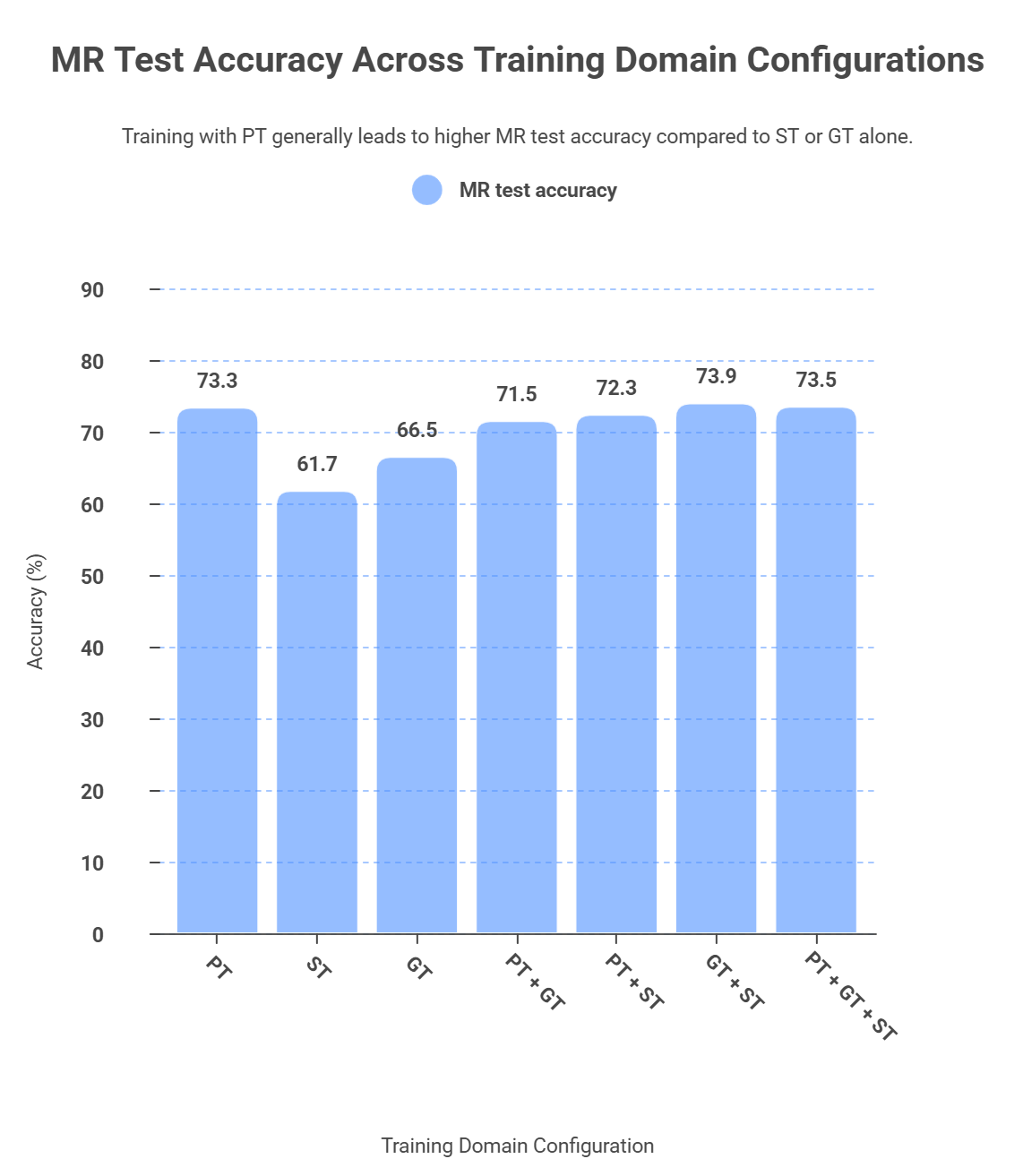}
\hfill
\includegraphics[width=0.48\textwidth]{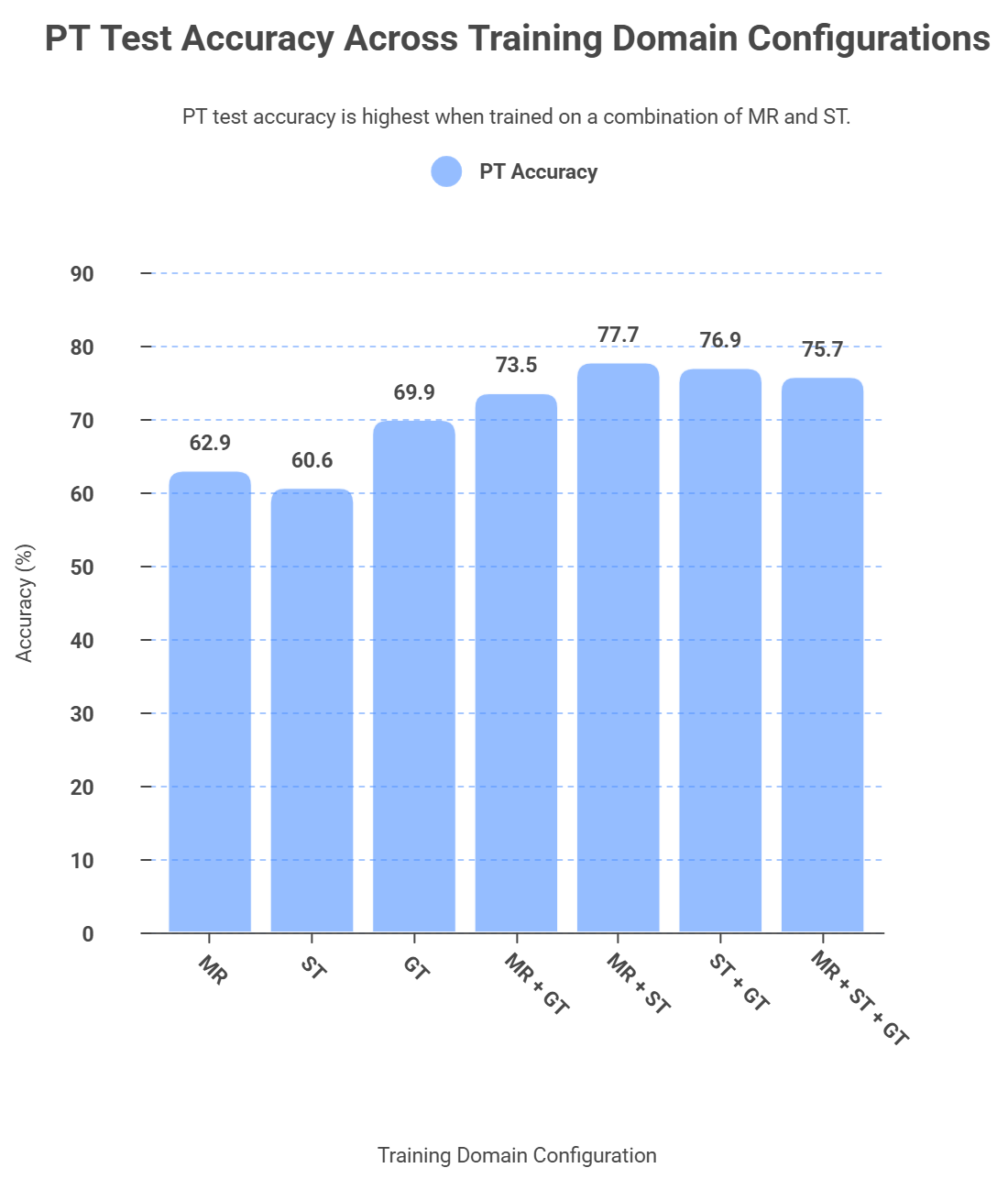}

\vspace{1.5cm}

\includegraphics[width=0.51\textwidth]{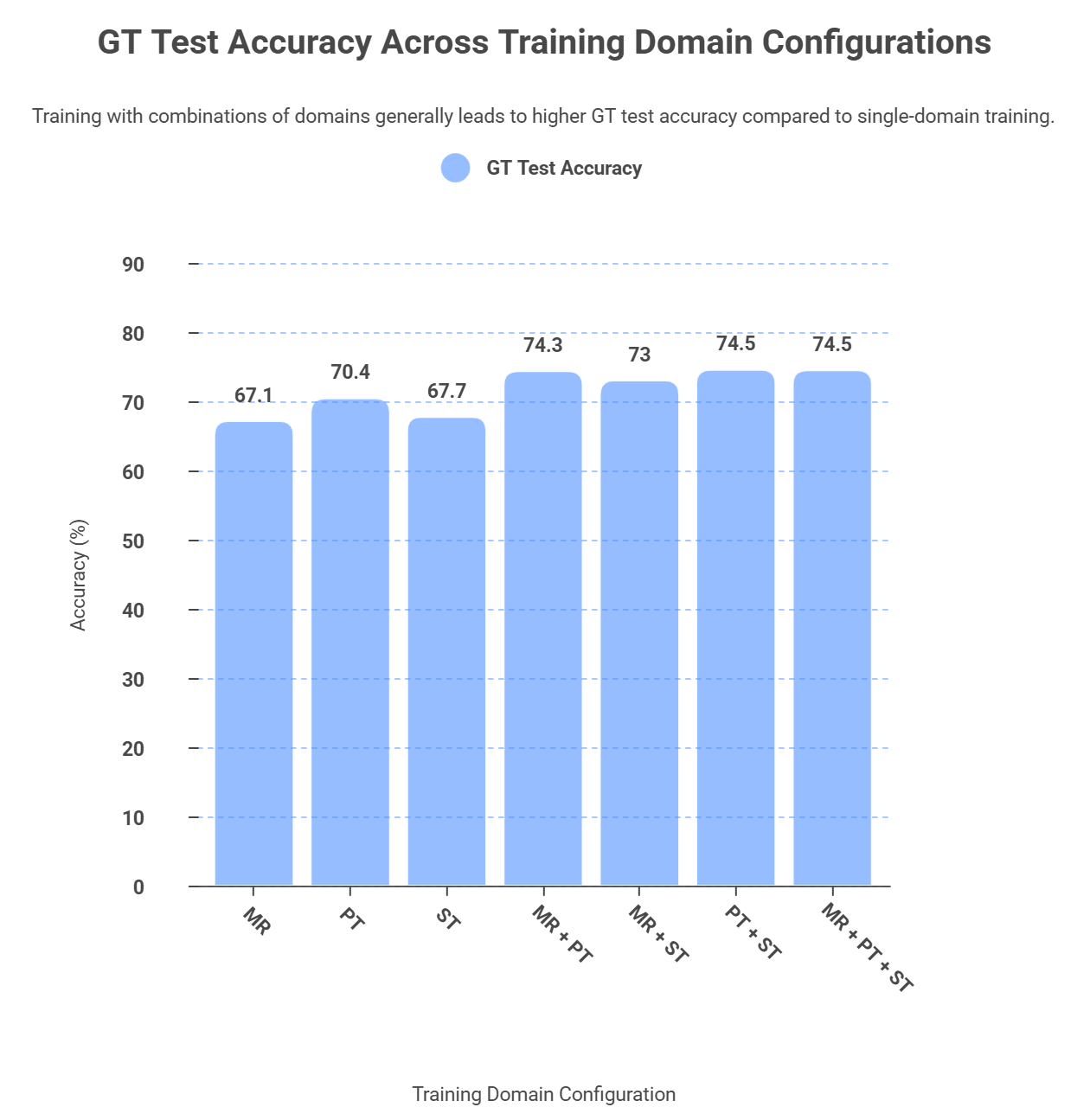}
\hfill
\includegraphics[width=0.45\textwidth]{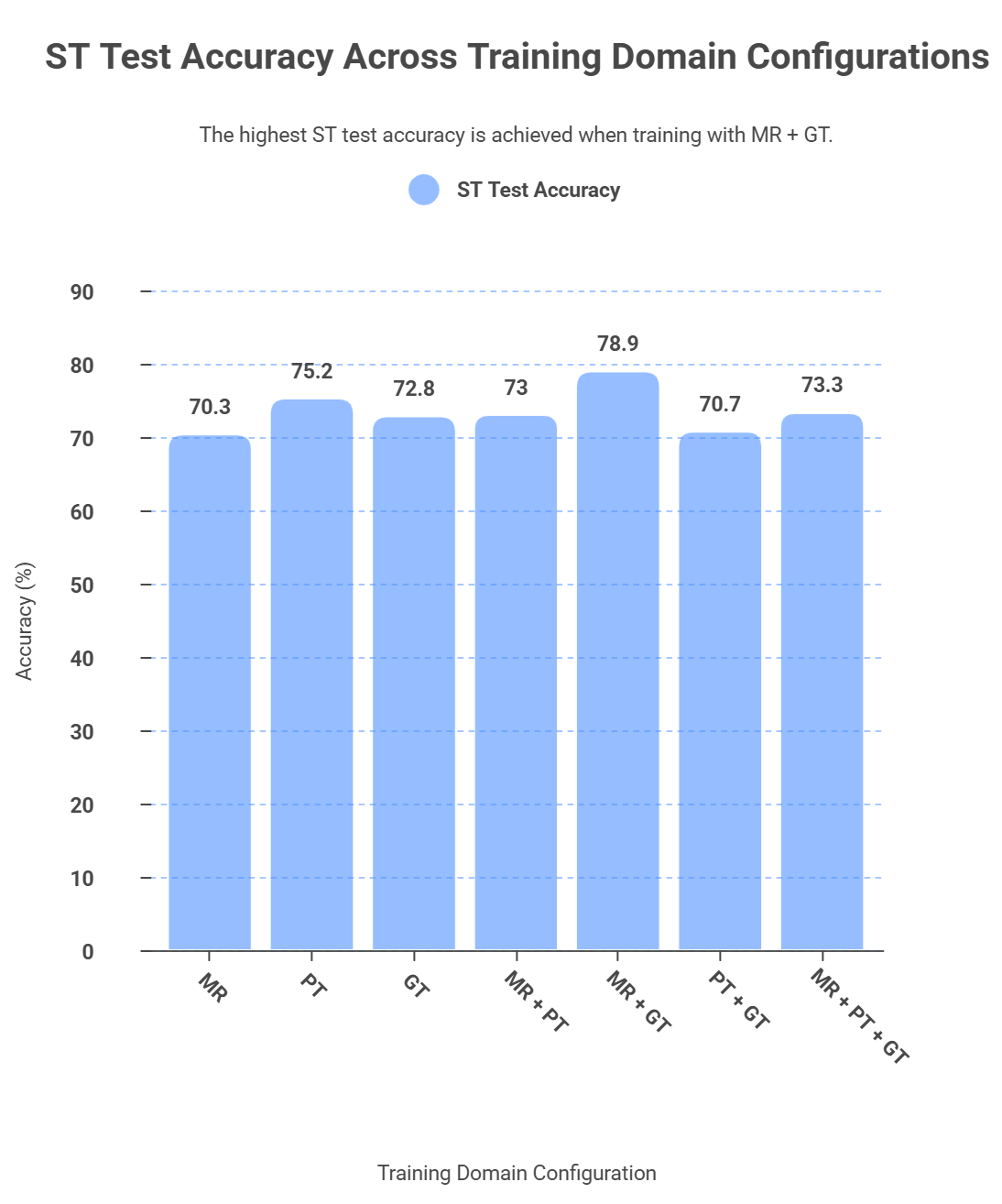}

\vspace*{\fill}
\caption{Cross-domain sentiment classification performance across different training configurations.}
\label{fig:cross_domain_results}
\end{figure*}

Let $x = (w_1, w_2, \dots, w_n)$ denote an input sentence composed of a sequence of tokens. Input text is tokenized using the MahaBERT-v2 tokenizer with a maximum sequence length of 128. Effective pre-processing and feature representation are critical for sentiment tasks, often bridging the performance gap between sentiment analysis and standard topic categorization \cite{haddi2013role,pingle2023augmentation}. Encoded representations, including input IDs and attention masks, serve as the primary features for model training.

\begin{equation}
Tokenizer(x) \rightarrow \{input\_ids, attention\_mask\}.
\end{equation}

These encoded representations serve as input features for model training.

Let $f_{\theta}$ denote the pretrained transformer encoder parameterized by $\theta$. For a given input sentence $x$, the encoder produces contextualized representations that capture semantic dependencies within the text. The pooled representation corresponding to the classification token is extracted as the sentence representation $h$. This representation is then passed through a linear classification layer defined as

\begin{equation}
z = Wh + b,
\end{equation}

where $W$ and $b$ denote the weight matrix and bias vector of the classifier. The predicted probability distribution over sentiment classes is obtained using the softmax function

\begin{equation}
\hat{y} = softmax(z).
\end{equation}

Model parameters are optimized by minimizing the cross-entropy loss

\begin{equation}
L = -\sum y \log(\hat{y}).
\end{equation}

Optimization is performed using the AdamW algorithm. The models are trained with a batch size of 16 for three epochs, and a weight decay factor of 0.01 is applied for regularization. The model achieving the highest validation accuracy during training is selected for final evaluation.

\subsection{Evaluation Protocol}

To assess cross-domain robustness, evaluation is conducted strictly on domains that are not included in the training subset $S$. This setup simulates a zero-shot generalization scenario in which the model encounters previously unseen domains during inference. In addition to overall evaluation, domain-specific performance is measured separately for each unseen domain $D_j \notin S$ in order to analyze domain-wise generalization behavior. This allows for a more detailed understanding of how the model performs across different domain distributions.

Model performance is evaluated using standard classification metrics including accuracy, precision, recall, and F1-score. To ensure balanced evaluation across sentiment classes, macro-averaged precision, recall, and F1-score are reported. This evaluation protocol enables systematic analysis of out-of-domain performance, zero-shot transfer capability, and the effect of increasing domain diversity during training. These metrics provide a comprehensive view of the model’s effectiveness in handling diverse and unseen data.

\section{Results and Discussion}


The experimental results highlight the impact of domain diversity on Marathi sentiment classification performance. Experiments were conducted on four domain-specific datasets, namely MahaSent-MR, MahaSent-PT, MahaSent-GT, and MahaSent-ST, under single-domain, two-domain, and three-domain training configurations. The task was formulated as a three-class sentiment classification problem, and models were evaluated across all target domains to analyze cross-domain generalization.

Results indicate that models trained on a single domain achieve strong in-domain performance but exhibit noticeable performance degradation when evaluated on unseen domains. For example, the model trained on MR achieved high accuracy on MR (78.53\%) but showed reduced accuracy on other domains such as GT and ST. Similarly, the PT-trained model performed best on PT (84.9\%) but demonstrated weaker cross-domain transfer. These observations suggest that single-domain training leads to domain-specific learning and limited robustness to distributional shifts.

In contrast, multi-domain training significantly improves generalization. Two-domain configurations such as MR+PT and MR+ST produced more balanced performance across domains, indicating that exposure to heterogeneous textual distributions enables the model to learn more transferable sentiment representations. In particular, incorporating conversational data such as ST improves adaptability to informal text.

\newpage
\begin{center}
\textbf{Table 2: Accuracies (\%) of Cross-Dataset Evaluation on Base Model}

\vspace{0.3cm}

\renewcommand{\arraystretch}{1.3}
\setlength{\tabcolsep}{8pt}

\resizebox{\columnwidth}{!}{
\begin{tabular}{lcccc}
\toprule
\textbf{Training Dataset} & \textbf{MR} & \textbf{PT} & \textbf{GT} & \textbf{ST} \\
\midrule
MR & 78.53 & 73.33 & 66.46 & 61.73 \\
PT & 62.93 & 84.90 & 69.87 & 60.60 \\
GT & 67.13 & 70.40 & 78.80 & 67.73 \\
ST & 70.33 & 75.24 & 72.80 & 79.07 \\
\bottomrule
\end{tabular}
}
\end{center}

\vspace{0.6cm}

\begin{center}
\textbf{Table 3: Accuracies of Combined Dataset (2 datasets) on Base Model}

\vspace{0.3cm}

\renewcommand{\arraystretch}{1.3}
\setlength{\tabcolsep}{8pt}

\resizebox{\columnwidth}{!}{
\begin{tabular}{lcccc}
\toprule
\textbf{Training Dataset} & \textbf{MR} & \textbf{PT} & \textbf{GT} & \textbf{ST} \\
\midrule
MR + PT & 81.67 & 85.24 & 74.33 & 73.00 \\
MR + GT & 80.80 & 73.51 & 75.93 & 78.93 \\
MR + ST & 80.93 & 77.69 & 73.00 & 79.47 \\
PT + GT & 71.46 & 85.06 & 79.00 & 70.73 \\
PT + ST & 72.33 & 86.00 & 74.53 & 79.60 \\
GT + ST & 73.93 & 76.93 & 79.73 & 80.26 \\
\bottomrule
\end{tabular}
}
\end{center}

\vspace{0.6cm}

\begin{center}
\textbf{Table 4: Accuracies (\%) of Combined Dataset (3 datasets) on Base Model}

\vspace{0.3cm}

\renewcommand{\arraystretch}{1.3}
\setlength{\tabcolsep}{8pt}

\resizebox{\columnwidth}{!}{
\begin{tabular}{lcccc}
\toprule
\textbf{Training Dataset} & \textbf{MR} & \textbf{PT} & \textbf{GT} & \textbf{ST} \\
\midrule
MR + PT + ST & 79.93 & 84.40 & 74.46 & 78.00 \\
MR + PT + GT & 79.86 & 84.40 & 78.06 & 73.26 \\
MR + ST + GT & 80.53 & 75.73 & 78.93 & 79.33 \\
PT + ST + GT & 73.46 & 85.50 & 79.46 & 79.73 \\
\bottomrule
\end{tabular}
}
\end{center}

\vspace{0.5cm}

Further improvements were observed in the three-domain configuration, where the models exhibited more consistent performance across all evaluation domains. Combinations such as MR+PT+ST and PT+ST+GT achieved relatively stable accuracies, suggesting that increased domain coverage helps the model capture more generalized sentiment cues. However, the performance gains over the two-domain configuration were comparatively smaller, indicating diminishing returns as additional domains are added.

Overall, the results demonstrate that domain diversity during training plays a critical role in improving cross-domain robustness for Marathi sentiment classification. While multi-domain training enhances generalization, the findings also suggest that selecting complementary domains is more beneficial than simply increasing the number of training domains.

\section{Conclusion}

In this paper, we examined the role of domain diversity in improving cross-domain generalization for Marathi sentiment classification using the MAHASENT-MD multi-domain framework. Experiments with single-domain, two-domain, and three-domain training configurations showed that models trained on a single domain achieve strong in-domain performance but struggle to generalize to unseen domains. Incorporating additional domains during training improves cross-domain robustness, with two-domain configurations providing notable performance gains. Although three-domain training further stabilizes performance across datasets, the improvements are relatively modest, indicating diminishing returns with increasing domain diversity.

Overall, the findings demonstrate that multi-domain training is an effective approach for developing robust Marathi sentiment classifiers in low-resource settings. The MAHASENT-MD framework provides insights for building scalable and domain-resilient sentiment analysis systems for Marathi and other low-resource languages.This highlights the importance of domain diversity.

\section*{Acknowledgments}

This work was conducted under the L3Cube-MahaNLP mentorship program. We sincerely thank our mentors and the L3Cube research team for their continuous guidance, support, and for providing the MAHASENT-MD and MahaCorpus datasets \cite{joshi2022mahanlp}, which were fundamental to this study. This research is part of the broader L3Cube-MahaNLP \cite{joshi2022mahanlp} project aimed at advancing Marathi Natural Language Processing.










\nocite{*}
\bibliography{main}

\end{document}